%% file: main.tex
\documentclass[11pt]{article}
\usepackage[utf8]{inputenc}
\usepackage{acl}

\usepackage{times}
\usepackage{latexsym}
\usepackage{multirow}
\usepackage{booktabs}
\usepackage{authblk}
\usepackage{tikz,todonotes,xspace}
\newcommand{\gr}[1]{\textsc{#1}}

\usepackage{microtype}

\title{One Model is All You Need: ByT5-Sanskrit, a Unified Model for Sanskrit NLP Tasks}

\author[1,3]{\bf Sebastian Nehrdich}
\author[2]{\bf Oliver Hellwig}
\author[3]{\bf Kurt Keutzer}

\affil[1]{Heinrich Heine University Düsseldorf\\}
\affil[2]{University of Zürich \\
Department of Comparative Language Science}
\affil[3]{University of California, Berkeley \\
Berkeley Artificial Intelligence Research (BAIR)}

\begin{document}
\maketitle
\begin{abstract}
 Morphologically rich languages are notoriously challenging to process for downstream NLP applications. This paper presents a new pretrained language model, ByT5-Sanskrit, designed for NLP applications involving the morphologically rich language Sanskrit. We evaluate ByT5-Sanskrit on established Sanskrit word segmentation tasks, where it outperforms previous data-driven approaches by a considerable margin and matches the performance of the current best lexicon-based model.
 It is easier to deploy and more robust to data not covered by external linguistic resources. It also achieves new state-of-the-art results in Vedic Sanskrit dependency parsing and OCR post-correction tasks. 
 Additionally, based on the Digital Corpus of Sanskrit, we introduce a novel multitask dataset for the joint training of Sanskrit word segmentation, lemmatization, and morphosyntactic tagging tasks.
 We fine-tune ByT5-Sanskrit on this dataset, creating a versatile multitask model for various downstream Sanskrit applications. 
 We have used this model in Sanskrit linguistic annotation projects, in information retrieval setups, and as a preprocessing step in a Sanskrit machine translation pipeline.
 We also show that our approach yields new best scores for lemmatization and dependency parsing of other morphologically rich languages.
 We thus demonstrate that byte-level pretrained language models can achieve excellent performance for morphologically rich languages, outperforming tokenizer-based models and presenting an important vector of exploration when constructing NLP pipelines for such languages.    
\end{abstract}

\section{Introduction}
It is generally acknowledged that morphologically rich languages (MRL) are challenging for NLP \citep{tsarfaty-etal-2020-spmrl}.
While language modeling has addressed this challenge, e.g. by integrating subword information \citep[see e.g.][]{bojanowski-etal-2017-enriching}, there is surprisingly little systematic research on how efficient models for low-level tasks such as tokenization, lemmatization, morphosyntactic analysis, and dependency parsing can be designed for MRLs. Access to this low-level information is relevant for downstream tasks such as information retrieval and question answering, as well as for linguistic and literary studies.

In this paper, we introduce a unified model that jointly performs these tasks for Sanskrit, an ancient South-Asian MRL, which has been continuously attested since 1,300 BCE. 
Vedic, its archaic level, primarily focusses on the description of the Soma and the fire sacrifice. 
Starting around 300 BCE, the majority of Sanskrit literature was composed in classical Sanskrit, encompassing a vast array of domains from religious hymns to scientific and narrative texts (see Table \ref{dataset}). 
Linguistic processing of Sanskrit poses challenges due to its rich morphology and vocabulary, free word order, heavy compounding, and particularly due to the phonetic merging of individual words into longer strings (Sandhi; see e.g. \citeauthor{gupta-etal-2020-evaluating} \citeyear{gupta-etal-2020-evaluating}), as can be observed in this example:
\newline
\begin{tabular}{llll}
\textit{yuvoḥ} & \textit{hi} & \textit{mātā} & \textit{aditiḥ} \\
your & indeed & mother & Aditi \\
\multicolumn{4}{l}{Aditi is indeed your mother.} \\
\multicolumn{4}{l}{With Sandhi: \textit{yuvorhi mātāditiḥ}} \\
\end{tabular}
\newline
Here, the words \textit{yuvoḥ} and \textit{hi} as well as \textit{mātā} and \textit{aditiḥ} are merged into longer strings, thereby changing their contact phonemes (\textit{ḥ}+\textit{h} $\rightarrow$ \textit{rh}, \textit{ā}+\textit{a} $\rightarrow$ \textit{ā}). While the synthesis of Sandhi is deterministic, its analysis is not, as the new phoneme \textit{-ā-} in \textit{māt\textbf{ā}ditiḥ} could also arise from \textit{a+a}, \textit{a+ā} or \textit{ā+ā}.
As a consequence, Sanskrit word segmentation (SWS) needs to be performed in order to enable tasks such as lemmatization, morphosyntactic tagging, and dependency parsing.

We propose a framework in which we pretrain a character-level Sanskrit language model based on ByT5 on a large body of Sanskrit data before jointly fine-tuning it on a number of downstream NLP tasks, which we reformulate as sequence-generation tasks. This paradigm brings large performance gains, leading to new SOTA results on established Sanskrit NLP benchmarks. We emphasize creating a system that is as simple as possible to train and deploy, without depending on complex pre- or postprocessing steps and retaining high performance on data that shows challenges such as OCR mistakes or the use of non-standard language not sufficiently covered by available linguistic resources.

In concrete terms, we achieve a \textbf{gain of 8.8 points on perfect sentence matching score (PM) for the Hackathon SWS benchmark} compared to the current state-of-the-art, 
while we come close by 0.13 in performance on the SIGHUM dataset to the currently best performing lexicon-driven model. \textbf{We achieve 4.88 points improvement on the SWS DCS 2018 benchmark}. On Vedic dependency parsing, we achieve \textbf{2.18 points improvement on UAS and 2.60 points on LAS} compared to the current state-of-the-art. On OCR post-correction, we outperform the currently best approach by \textbf{0.29 lower CER and 3.16 lower WER}. We also show that our approach yields the best performance on lemmatization and dependency parsing for three other MLR languages.

We also present a novel dataset for the training and evaluation of three central Sanskrit NLP tasks based on the Digital Corpus of Sanskrit (DCS): Word segmentation, lemmatization, and morphosyntactic tagging. We show that our pretrained model outperforms other baselines on these new tasks. We also demonstrate that jointly training on the tasks of SWS, lemmatization, and morphosyntactic tagging on top of the pretrained language model leads to the best performance. This enables the deployment of one single model without dependence on external linguistic resources to handle all relevant NLP tasks for annotated Sanskrit corpus building with the best performance. We show that training and evaluating this model on pseudo-paragraph-level, where multiple sentences are predicted at once, gives a distinct performance advantage due to the available contextual information. 

In Section \ref{related-research}, we give an overview of the relevant research literature. In Section \ref{datasets}, we discuss the pretraining and fine-tuning datasets used in this paper. Section \ref{proposed-method} introduces the layout of our proposed multitask framework.
In Section \ref{experiments}, we first evaluate the model on established Sanskrit word segmentation, Vedic Sanskrit dependency parsing, OCR post-correction tasks, as well as on other MLR languages, and then present the performance of the unified model trained on the new dataset. We also perform a detailed manual analysis of the error patterns of the multitask model.
We make the code, all relevant datasets, the pretrained base model as well as the fine-tuned multitask model available under the Apache license 2.0 at \url{https://github.com/sebastian-nehrdich/byt5-sanskrit-analyzers}.


\section{Related Research}\label{related-research}
The pretrain-fine-tune paradigm, where a pretrained language model (PLM) trained on a large corpus of unlabeled data is subsequently fine-tuned on a smaller dataset of task-specific labeled data, is the de-facto standard approach for NLP tasks such as part-of-speech and morphosyntactic tagging, sentence classification, and many more since the publication of the encoder-only approaches BERT \cite{bert} and ELMo \cite{elmo} in 2018.
When it comes to morphologically rich languages, the good performance of this paradigm is demonstrated for Turkish in \citet{ozcift-turkish}, while \citet{latinbert} and \citet{nehrdich-hellwig-2022-latin} show the superior performance of BERT on linguistic annotation tasks for the morphologically rich classical language Latin.  

T5 \cite{t5paper} introduced a new pretraining paradigm where both encoder and decoder are trained. 
This encoder-decoder architecture enables the fine-tuning of the same base model on diverse tasks such as translation, question answering, and text classification with the same hyperparameters and loss function. \citet{t0paper} further show how the T5 paradigm can be used efficiently in a multitask setup with large variation between the different tasks.
For morphologically rich languages, language models that make use of character-level information show superior performance to those operating on word-level alone \cite{gerz2018}. While a number of openly available pretrained language models exist, only \citet{byt5paper} followed a tokenizer-free byte-level approach, resulting in strong performance on linguistic tasks and achieving the best performance on the morphological inflection task. 

Most approaches to Sanskrit NLP tasks such as Sanskrit word segmentation (SWS) can be broadly separated into two groups: lexicon-based and data-driven. For a recent, comprehensive overview of the relevant literature, see \citet{sandhan2022}. Lexicon-driven approaches rely on external linguistic resources, while data-driven approaches learn from data alone and are therefore less complex to train and deploy. The main drawback of data-driven approaches is that they cannot access latent knowledge contained in lexical resources.  
\citet{sandhan2022} combine lexicon-based and data-driven aspects, formulating SWS as a character-level sequence labeling task that uses lexical information whenever available. \citet{krishna2020} presents a lexicon-based multitask model that handles SWS, morphological parsing, dependency parsing, syntactic linearization, and prosodic linearization. To our knowledge, this is the only other published multitask approach to central Sanskrit NLP tasks. 

Pretrained language models supporting Sanskrit are available, but they are not yet widely used for Sanskrit linguistic tasks. \citet{xlm2019} included Sanskrit data in its pretraining setup. \citet{hellwig2023} trained and evaluated encoder-only PLMs for the task of Vedic Sanskrit dependency parsing, coming to the conclusion that they do not offer clear advantages in performance yet due to the comparatively small amount of training data used. 

\section{Data}\label{datasets}
For pretraining, we use the Sanskrit data of the Sangraha dataset  \cite{khan2024indicllmsuite} as a basis, which mostly consists of data gained by a comprehensive OCR effort of the Sanskrit-related literature available at the Internet Archive\footnote{\url{archive.org}}. We only use the language-verified split of this dataset and none of the synthetic data.
We decided to use this noisy OCR-based dataset following the observation made in \citet{latinbert}, where a PLM for Latin trained on a noisy corpus consisting of largely OCR'd data achieved new state of the art results on Latin POS tagging tasks.
We augment this data with high-quality human input Sanskrit data from the GRETIL collection \footnote{\url{https://gretil.sub.uni-goettingen.de/gretil.html}} and the Digital Sanskrit Buddhist Canon.\footnote{\url{https://www.dsbcproject.org/}} The statistics of the dataset are shown in Table \ref{dataset-pretraining}.

We use IAST transliteration for pretraining as well as all of the fine-tuning tasks, as this yields clear efficiency advantages compared to Devanagari when training on the individual byte level, with half the bytes needed. While other transliteration schemes such as SLP1 offer further small gains in efficiency, we decided against using them as the human readability advantages of IAST lead to less overhead during training and evaluation, as well as less complex deployment pipelines.  
\input{input/dataset-pretraining}

\subsection{Fine-tuning Dataset}\label{section-dataset}
The fine-tuning data utilized in this study for the SWS, lemmatization, morphological tagging, and dependency parsing tasks comes from the Digital Corpus of Sanskrit (DCS; \citeauthor{hellwig_DCS} \citeyear{hellwig_DCS}), a collection of classical and Vedic texts with manually validated lexical and morphosyntactic annotations. For some Vedic texts, the DCS also provides manually validated syntactic annotations \citep{hellwig2023}. The complete annotation is available as text files in CoNLL-U format,\footnote{\url{https://github.com/OliverHellwig/sanskrit/tree/master/dcs/data/conllu}} serving as input for the multitask and dependency parsing models described in this paper. We use a snapshot of the DCS dataset from April 2024. 
Table \ref{dataset} gives an overview of the DCS fine-tuning data, showing its bias towards narrative (epics, Pur\=a\d nas), Vedic, and scientific texts.
\input{input/dataset}

\section{Proposed Method}\label{proposed-method}
\begin{figure*}[htbp]
  \centering
  \includegraphics[scale=0.35]{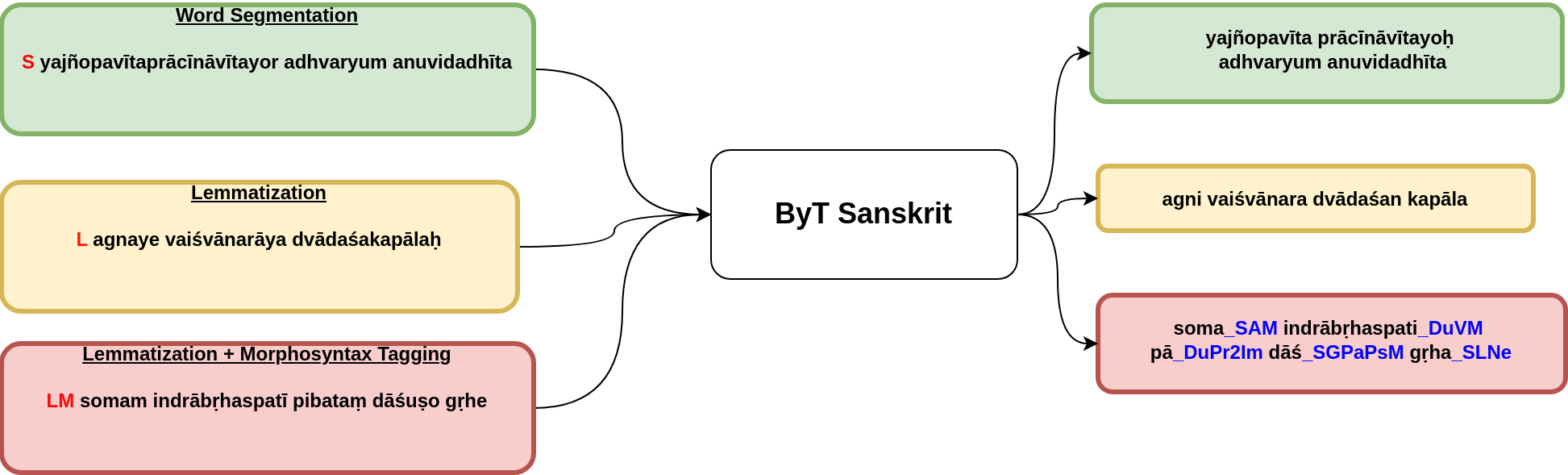}
  \caption{Sanskrit Multitask Formulation: All tasks are converted into sequence-generation tasks. For each task, we prepend prompt tokens (S, L, LM, here marked in red) in order to enable the model to distinguish between tasks. For efficient training and inference, we use a novel serialization strategy to compress the morphosyntactic tags into as few characters as possible, here marked in blue.}
  \label{schema-diagram}
\end{figure*}

We propose the combination of the following paradigms in order to generate an efficient, high-performing end-to-end framework for various Sanskrit NLP tasks:
We first pretrain a byte-level Sanskrit PLM based on the ByT5 architecture, which is distributed under the Apache license 2.0, overcoming the limitation of lack of access to latent information for data-driven approaches. Then, we reformulate the central Sanskrit NLP tasks of word segmentation, lemmatization, and morphosyntactic tagging as sequence generation tasks, using a novel serialization strategy. 
In order to distinguish between the different tasks,  we use prefix letters at the beginning of the input sequence to indicate the task. ``S'' for segmentation, ``L'' for lemmatization, and ``M'' for morphosyntactic tagging.
Inspired by T0 \cite{t0paper}, we combine these tasks into a unified multitask setup, enabling the fine-tuning of a single model to handle all of them simultaneously. The schema of this approach is demonstrated in Figure \ref{schema-diagram}.

The full morphosyntactic tags of the DCS consume on average 46 characters, making their prediction with a byte-level LM challenging. We therefore propose a serialization strategy by manually mapping the morphosyntactic tags to unused letter combinations of the IAST alphabet, reducing the number of needed tokens per tag significantly. The full tags can be restored based on this mapping without information loss. Figure \ref{serialization} demonstrates this process.
On average, the compression ratio of this method is 0.14. 
\input{input/serialization}  

\section{Experiments}\label{experiments}
Models were trained on GPU nodes of 8 NVIDIA
A6000 48GB GPUs. The time needed for pretraining was one week, while the fine-tuning runs varied between 2 and 8 hours. The joint multitask model took 32 hours to finetune. 
We leveraged the DeepSpeed library
\url{https://www.deepspeed.ai/} for training in half precision bf16, and for making efficient use of the multi-GPU setup.

For the foundation model, we further pretrain a ByT5 model \citep{byt5paper} in the ``base'' configuration with 582M parameters on the entire dataset for 100,000 steps with a batch size of 512 and a sequence length of 512. The resulting model is called ByT5-Sanskrit in this paper. According to the scaling laws presented in \citet{hoffmann2022}, the optimal number of parameters for our training dataset size of 6.5B tokens is about 325M parameters. This matches the ByT5 ``small'' configuration with 300M parameters.  We decided to train a model one category larger than that to ensure we get optimal performance. 

\subsection{Evaluation on Previous Sanskrit Word Segmentation Tasks}
In order to examine how ByT5-Sanskrit performs in comparison to other baselines, we fine-tune it on a selection of different previously established Sanskrit word segmentation tasks, each of which used its own dataset. 

The SIGHUM and Hackathon datasets are adapted from \citet{sandhan2022}. DCS 2018 is the dataset presented in \citet{hellwig2018}. rcNN-SS denotes a character-based segmentation algorithm that performs joint compound and Sandhi splitting using a combination of recurrent and convolutional operations \citep{hellwig2018}. 
TransLIST is the model described in \citet{sandhan2022}, which uses a combination of character-level and lexicon-based word input with a transformer model. 

The results of our comparison are shown in Table \ref{results-sws}. Since TransLIST, due to its elaborate preprocessing pipeline, is not compatible with the DCS 2018 dataset, we cannot evaluate it in that setting. On DSC 2018 and Hackathon, ByT5-Sanskrit outperforms the existing best baselines with a very considerable margin, while it comes close to the best-performing lexicon-based model, TransLIST, on the SIGHUM dataset. The results show that ByT5-Sanskrit successfully learns latent features of the Sanskrit language and achieves very strong performance without relying on lexical resources. The performance gain on the Hackathon task is especially noteworthy, as this task has the smallest train split of the three with 89k samples, indicating that fine-tuning ByT5-Sanskrit is very sample efficient. 

Compared to ByT5-Sanskrit, TransLIST shows more variation in performance between SIGHUM and Hackathon, indicating that the quality of data preprocessing determines the quality of the outcome for TransLIST to a significant degree. ByT5-Sanskrit on the other hand shows consistent performance improvements on all three tasks, making elaborate lexical pre-processing unnecessary to reach competitive performance.

\begin{table}
\resizebox{ \columnwidth}{!}{%
    \begin{tabular}{l|c|c|c}
 & DCS 2018 & SIGHUM & Hackathon \\
Model & (509k samples) & (99k samples) & (89k samples) \\
    \hline
         \input{input/results-sandhi-split-older-datasets}

    \end{tabular}
    }
    \caption{\label{results-sws}{Sentence level perfect matches on previous Sanskrit word segmentation tasks. Results for rcNN-SS and TransLIST are reported based on the respective publications. Due to data incompatibility, we cannot evaluate TransLIST on the DCS 2018 task. Size of training dataset in parentheses.}}
\end{table}

\subsection{Vedic Dependency Parsing}
We also evaluate the performance of ByT5-Sanskrit on the Vedic Sanskrit dependency parsing task. We follow the serialization strategy of \citet{Lin2022DependencyPV} and reformulate dependency parsing as a sequence generation task. We compare our results against the biaffine architecture \citep{dozat2016} in its best performing configuration as presented in \citet{hellwig2023}. Using the latest version of the dependency annotated data of the DCS for our experiments,
we extract a total number of 24,807 sentences with gold dependency, part of speech, and morphosyntactic annotation. We use 90\% for training, and 5\% for each evaluation and testing. 
Following the setup in \citet{hellwig2023}, we exclude \d Rgvedic data from the test and evaluation split and apply the augmentation strategy of randomly concatenating up to four 
sentences from the training set. 
Moreover, we replace the POS and morphosyntactic information of old Vedic citations (mantras) with a special 
tag. 
The biaffine model and ByT5-Sanskrit are trained and evaluated on the same data. We use 50 epochs for training. We evaluate both models in two settings: One without any additional linguistic information, using only the surface form of a word (None), and one where all available linguistic features (POS tags, morphosyntax, punctuation) are used (All). 
\begin{table}
    \begin{tabular}{l|c|c|c|c}
  \multicolumn{1}{c}{} & \multicolumn{2}{c|}{Biaffine} & \multicolumn{2}{c}{ByT5-Sanskrit} \\
Setting & UAS & LAS & UAS & LAS \\
    \hline
         \input{input/results-dependency-parsing}
    \end{tabular}
    \caption{\label{results-dp}{UAS and LAS for the  Vedic dependency parsing. ``None'': only surface forms used; 
    ``ALL'': all linguistic gold information used 
    }}
\end{table}

The results in Table \ref{results-dp} show significant performance improvements of 2.18 in UAS and 2.60 in LAS over the biaffine baseline. Especially noteworthy is the observation that the ByT5-Sanskrit-based parser without any additional linguistic information comes close to the performance of the biaffine parser with support of gold data. These results are in line with the observations made in \citet{nehrdich-hellwig-2022-latin}, where the addition of a strong Latin PLM boosted dependency parsing performance very significantly on three different Latin treebanks, with configurations based on the Latin PLM alone matching those that make use of gold annotation without the PLM. 

\subsection{Sanskrit OCR Post-correction}
We also evaluate our model on the task of Sanskrit OCR post-correction as defined in \citet{sanskritocr2022}. Our results are presented in Table \ref{results-ocr}. We fine-tune ByT5-Sanskrit with a sequence length of 512. The results show that ByT5-Sanskrit also achieves the best performance on this task. 

\subsection{Lemmatization and Dependency Parsing on other MLR Languages}
In order to test whether our proposed framework generalizes to other MLR languages, we conduct experiments on lemmatization and dependency parsing for three MLR languages: Bulgarian, Romanian, and Turkish. The data is taken from the Universal Dependency \cite{nivre2016} 2.2 release. As the base model for finetuning, we use ByT5 in the "base" configuration without further pretraining. We show the results in Table \ref{results-other-languages}. Since our serialization strategy requires language expertise, we cannot evaluate our framework on morphosyntactic tagging for these languages. We compare our approach against the current best baseline UDPipe \cite{straka2019} for lemmatization, for dependency parsing we also compare against DPSG \cite{Lin2022DependencyPV}, since their approach reaches the currently best results on these languages and is structurally very similar to our, with the main difference being that we use a byte-level PLM, while they use the tokenizer-based PLM mT5.  The results show that our approach outperforms the previous best baselines on lemmatization for two languages, while outperforming the previous baselines on dependency parsing for all languages. This shows that the performance advantages of byte-level PLMs generalize to other morphologically rich languages.
\begin{table}
\small
    \begin{tabular}{l|c|c}
  \multicolumn{1}{c}{} & Lemma & Dep. Parsing \\
Language & Acc & LAS \\
    \hline
         \input{input/results-other-languages}
    \end{tabular}
    \caption{\label{results-other-languages}{Lemmatization and dependency parsing results on three other MLR languages based on ByT5 base. The UDPipe results are reported based on \citet{straka2019}, DPSG based on \citet{Lin2022DependencyPV}. }}
\end{table}

\begin{table}
\centering
    \begin{tabular}{l|c|c}
  Model & CER & WER \\ 
    \hline
         \input{input/results-ocr}
    \end{tabular}
    \caption{\label{results-ocr}{CER and WER results for the Sanskrit OCR post-correction task. ByT5-Small are the results as presented in \citet{sanskritocr2022}.}}
\end{table}

\subsection{Joint Sanskrit Word Segmentation, Lemmatization and Morpho-syntax Tagging Task}\label{sec-joint}

We use a snapshot of the DCS from April 2024 as the basis for our experiments with a total number of 601,403 
sentences. We hold back 8,190 sentences for evaluation and 8,398 sentences for testing. We keep the original order of the sentences, ensuring that this data can also be used to train models on longer sections of text. 
\label{reconstructed-forms}The DCS presents a challenge for our word segmentation model because the forms without Sandhi were not consistently recorded during the initial annotation of the DCS. This incomplete annotation affects 65.8\% of all words in the DCS, primarily from classical Sanskrit, whereas unsandhied forms are recorded for most Vedic and some Buddhist texts. When generating conllu files from the DCS, a heuristic that supplements missing unsandhied forms is employed to address this inconsistency. While this heuristic can occasionally produce morphologically correct but unattested nominal forms (e.g., generating \textit{hṛdayataḥ} instead of the attested \textit{hṛdayāt} for the ablative singular of \textit{hṛdaya-} `heart'), a cursory examination suggests such cases are infrequent (1-3\% of all words).
Since we believe that the heuristically generated forms can nonetheless be useful for the training, we include them in the training set and prepend a special flag ``R'' at the beginning of each line 
containing such forms. The test and validation splits do not contain any reconstructed forms and \label{test-data-bias}are therefore strongly biased towards Vedic texts. This makes the annotation tasks more challenging because Vedic texts are underrepresented in the DCS (see Table \ref{dataset}).

 Since the tasks can be combined arbitrarily, we decided to limit the experiments to a number of settings with real-world relevance. We provide the data on sentence- and pseudo-paragraph level. Pseudo-paragraphs are constructed by concatenating adjunct sentences with a length of up to 512 characters, giving the model the possibility to utilize contextual information.

We show the results in Table \ref{results-all}. As the lexical resources used for \citet{sandhan2022} and \citet{krishna2020} are not compatible with our dataset, we cannot evaluate their performance in this setting. Due to resource constraints, we could not evaluate mT5 on paragraph level. 
ByT-Sanskrit outperforms all other models on these tasks. The visible gains compared to ByT5 indicate that even in a multitask setup with a large fine-tuning dataset, the prior knowledge from the pretraining stage brings distinct performance advantages. The weaker performance of mT5 shows that tokenizer-based models don't perform as well in this setting. All models perform better on pseudo-paragraph level, showing that contextual information beyond sentence boundaries is crucial for Sanskrit linguistic tasks and should be used wherever possible. The visible performance drop that occurs when including morphosyntactic tagging can be explained by the fact that these tags are often ambiguous, as will be discussed in the error analysis below. 
\begin{table}
\centering 
\small
\setlength{\tabcolsep}{5pt}
    \begin{tabular}{l|c|c|c|c|c}
 \multicolumn{1}{c}{} & mT5 & \multicolumn{2}{c|}{ByT5} & \multicolumn{2}{c}{ByT5-Sanskrit} \\
Task & Sen & Sen & Par & Sen & Par \\
    \hline
         \input{input/results-new-dataset}
    \end{tabular}
    \caption{Sentence level perfect match results for the multitask experiment. "S" denotes the task of Sanskrit word segmentation, "L" the task of lemmatization, and "M" the task of morphosyntax tagging.}
    \label{results-all}
\end{table}


\begin{figure}
    \centering
    \includegraphics[width=0.4\textwidth]{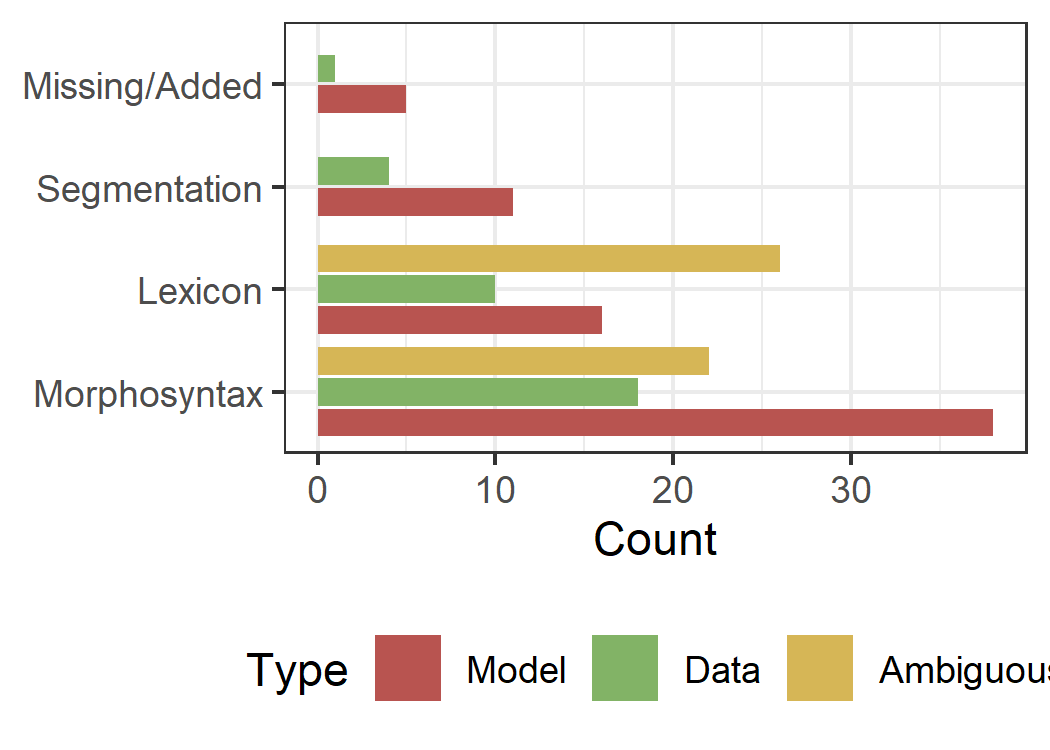}
    \caption{Results of the detailed error analysis }
    \label{fig-error-analysis}
\end{figure}

\subsection{Error analysis}
For a detailed error analysis, one author of this paper inspected 150 randomly drawn sentences in which the model result differs from ground truth. The differences were categorized into three classes: (1) errors in the ground truth data which were corrected by the model; (2) ambiguous cases where both ground truth and model result are acceptable; and (3) model errors. These three classes were further subsetted with fine-grained error labels. Most basically, the segmentation of a string may have failed or parts of the analysis is missing or added. Figure \ref{fig-error-analysis} shows that these cases 
constitute only a small part of all errors. Notably, missegmentations as well as missing words are also present in the ground truth, e.g. when a meaningful linguistic analysis of a string 
was impossible.

Most differences are observed at the lexical and morphological levels. The pattern of model errors aligns with the summary in \citet{gupta-etal-2020-evaluating}, primarily relating to nominal endings that denote more than one case. For instance, an error is seen in the phrase \textit{uttamāyā diśaḥ} `from the highest region', where both the correctly reconstructed ending \textit{-āyāḥ} in \textit{uttamāyāḥ} and \textit{-aḥ} in \textit{diśaḥ} can signify a genitive or ablative singular. However, the sentential context of this phrase unambiguously indicates an ablative interpretation.
Errors in the analysis of verbal morphology are less common and typically occur with unusual and rare forms. An example is \textit{āvarvratataḥ}, the genitive singular of a participle derived from the intensive of the verb \textit{ā vart-} `revolve' (`of someone who rotates intensively'). The model misinterpreted this complex form as the nominative singular of a newly coined noun \textit{āvarvrata-}.

Both gold and silver exhibit a significant number of ambiguities. 
Consider the following phrase:\newline
\begin{tabular}{ll}
	 \textit{apareṇa} & \textit{śālāyāḥ} \\
	western-\gr{Adj}.\gr{Ins}.\gr{Sg} & hall-\gr{Gen/Abl}.\gr{Sg} \\
    to\_the\_west\_of-\gr{Prep} & \\
	\multicolumn{2}{l}{`to the west of the hall'} \\
\end{tabular}
\newline
Here, \textit{apareṇa} is an example of a lexical ambiguity. While, at the level of morphosyntax, the word is the instrumental singular of the adjective \textit{apara-} `western', it can be argued that this word 
became grammaticalized in Vedic, as was shown for Vedic \textit{madhye} `in the middle' $>$ Hindi \textit{meṃ} `in' \citep{reinoehl-config}. Although the grammaticalized reading is preferred in the DCS, analyzing \textit{apareṇa} as an inflected adjective is perfectly valid, given our limited knowledge about the temporal dynamics of grammaticalization processes. Similarly, while the genitive reading of \textit{śālāyāḥ} is the preferred analysis in the DCS, the ablative cannot be ruled out here, leading to a morphological ambiguity.

Beyond this, lexical ambiguities in the DCS primarily arise from compound splitting. The DCS follows major Sanskrit dictionaries in not splitting compounds deemed to have non-compositional meanings. 
Given the lexical transparency of most Sanskrit compounds, splitting them into their constituent parts is often a justifiable approach. For instance, the DCS keeps the compound \textit{ādikaraḥ} ``creator'' intact, but our model reasonably splits it into \textit{ādi-karaḥ} ``beginning-maker''.

In about one third of all cases, our model corrects a wrong analysis in the DCS. 
One case is \textit{dvādaśa-kapālam} `(consisting of) twelve cups' where the first word is the compound form of the numeral \textit{dvādaśan-} `twelve', but not, as recorded in the DCS, of the adjective \textit{dvādaśa-} `twelfth'. 
Apparently our model achieves a quality high enough to be usable for error detection in the ground truth data. While retraining ByT5-Sanskrit with corrected data is not likely to improve its quality, such error correction may nevertheless be useful for linguistic studies. 

Overall, more than half of the 150 sentences inspected (80 or 53.3\%) revealed errors in the source data or alternative valid readings. Together with the bias in the test data (see Section \ref{test-data-bias}), this result indicates that the predictive quality of our model is higher than indicated by the numbers in Table \ref{results-all}.

\subsection{Ablation Study}
To assess the impact of joint training on multiple tasks, we conducted an ablation study where we fine-tuned ByT5-Sanskrit on selected tasks individually. The results, presented in the upper half of Table \ref{ablation}, clearly demonstrate that individual task training diminishes performance for both segmentation and lemmatization tasks. This confirms that transfer learning across different tasks contributes to enhanced overall performance.

A second experiment evaluated the effect of removing samples containing reconstructed surface forms (refer to Section \ref{reconstructed-forms}). This condition reduced the training sample size to merely 26.22\% of the original data, effectively serving as an ablation experiment probing the dataset size. Despite a noticeable negative impact on performance (as seen in the lower half of Table \ref{ablation}), the effect was less pronounced than we anticipated. We hypothesize that this behavior can primarily be attributed to the strong priors of the ByT5-Sanskrit model. Concurrently, removing sentences with reconstructed forms from training rendered the distributions of training and test data, which exclusively contain such sentences, more similar. This suggests that our approach is viable even for languages with a limited amount of labeled training data.


\begin{table}
    \begin{tabular}{l|c}
  Task & Sentence PM \\ 
    \hline
         \input{input/ablation}
    \end{tabular}
    \caption{\label{ablation}{Ablation Study where we fine-tune ByT-Sanskrit on individual tasks seperately to show the performance difference to the multitask setup. W/o rec. indicates the setting where reconstructed forms (see Section \ref{sec-joint}) are not used in the training dataset. Results are given in sentence level perfect matches.}}
\end{table}

\section{Conclusion and Future Work}
We have demonstrated that by pretraining a byte-level language model on a large collection of mostly noisy data, new state-of-the-art results for Sanskrit word segmentation, Vedic dependency parsing, and OCR post-correction are achieved, closing the performance gap between lexicon-based and data-driven Sanskrit NLP approaches. We further demonstrated that this pretrained language model can be used as a basis for a multitask model that handles word segmentation, lemmatization, and morphosyntactic tagging jointly with high accuracy. We further demonstrated that this multitask model benefits greatly from training and inference on pseudo-paragraph-level.
For the joint fine-tuning on these tasks, we presented a novel dataset. The resulting unified model, being independent of external linguistic resources, is simple to deploy and is already used for Sanskrit corpus annotation projects as well as in information retrieval and machine translation setups.
We also showed that our approach generalizes to other morphologically rich languages, where the application of a byte-level PLM yields best results for two languages on lemmatization and for three languages on dependency parsing. We thus establish that byte-level PLMs are a crucial vector of exploration when builiding NLP pipelines for MLR languages.

\section{Limitations}
Our model currently does not adequately address the homonymy of words. In the DCS, 7.5\% of all lemmata, or tokenized words, have at least one homonym. These homonyms account for a significant 57.5\% of all tokenized words. However, this percentage is somewhat misleading. The primary contributors to this high rate of homonymy are indeclinable words such as \textit{ca} `and' and \textit{iti} `thus’. These words are used as nouns in grammatical literature, most notably in Pāṇini’s Aṣṭādhyāyī, where their case endings indicate grammatical uses (e.g., at Aṣṭādhyāyī 1.1.16: \textit{… itau anārṣe} `in front of (the particle) \textit{iti} in non-Vedic texts’). In non-grammatical texts, these words almost always have their non-technical meaning. Similar considerations apply to the use of nominalized verbal roots in grammatical texts.

There are more problematic, but less frequent cases. For instance, the word \textit{veda} has four different lemmata recorded in the DCS: (1) the famous text collection of the same name, (2) `finding, obtaining’, (3) a small broom, and (4) the name of a man. At least homonyms (1) and (3) are regularly attested in Vedic and classical Sanskrit. Merging them into one lemma is lexicographically inadequate: while (1) and (2) may be etymologically related, (3) and probably also (4) are not \citep[see][579-581]{mayrhofer-ewia-1}. However, the context of their occurrence typically indicates very clearly which of the lemmata is meant.

To address this issue, we plan to mark lemmata with homonyms by numeric affixes in future versions of our model.

\bibliography{anthology}

\end{document}

%% file: input/dataset-pretraining.tex
\begin{table}
\centering
\begin{tabular}{lr}
\toprule
Source & Number of Characters \\
\midrule
IndicLLMSuite & 5,173,251,798 \\
GRETIL & 253,712,457\\
DSBC & 2,473,226\\
\bottomrule
\end{tabular}
\caption{\label{dataset-pretraining} Composition of the pretraining dataset.  Number of characters is measured in character count in IAST roman transliteration.}
\end{table}

%% file: input/dataset.tex
\begin{table}

\centering
\begin{tabular}{lr}
\toprule

Category & Number of Characters \\
\midrule
Epics & 9,814,868 \\ 
Vedic & 7,211,586 \\ 
Science & 6,299,576 \\ 
Purāṇa & 4,682,010 \\ 
Poetry & 2,028,535 \\ 
Buddhist & 1,762,012 \\ 
other & 2,728,511 \\
\bottomrule
\end{tabular}
\caption{Distribution of the fine-tuning data according to different categories. Number of characters is measured in character count in IAST roman transliteration.}
\label{dataset}
\end{table}

%% file: input/serialization.tex
\begin{figure*}[htbp]
    \centering
   \includegraphics[scale=0.40]{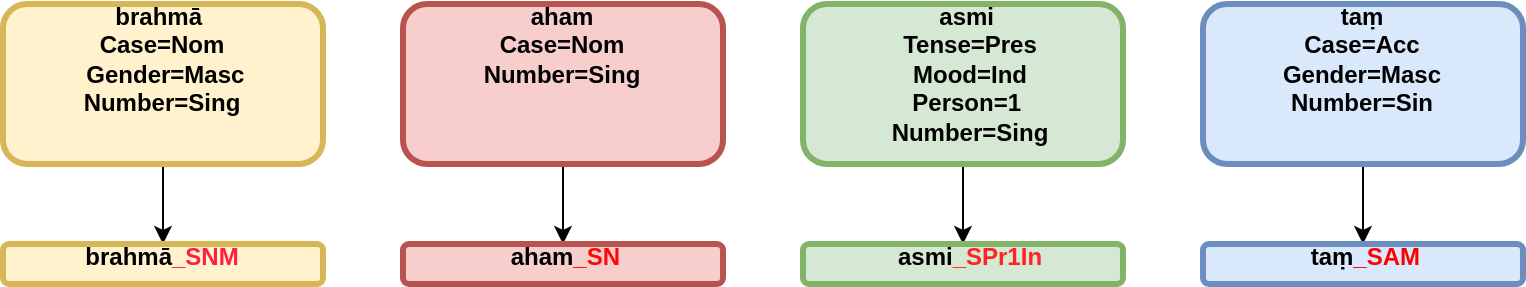}
    \caption{\label{serialization}Serialization for the morphosyntactic tagging task. The abbreviated tags are highlighted in red. We use spaces as separation token between words.}
\end{figure*}

%% file: input/results-sandhi-split-older-datasets.tex
rcNN-SS & 85.2 & 87.08 & 77.62 \\
TransLIST & - & \textbf{93.97} & 85.47 \\
ByT5-Sanskrit & \textbf{90.11} & 93.83 & \textbf{94.29} \\

%% file: input/results-dependency-parsing.tex
None & 77.68 & 70.67 & \textbf{86.54} & \textbf{81.54} \\
ALL & 86.86 & 81.98 & \textbf{89.04} & \textbf{84.58} 

%% file: input/results-other-languages.tex
Turkish IMST (UDPipe) & 96.01 & 67.56 \\
Turkish IMST (Ours, ByT5) & \textbf{97.94} & \textbf{77.00} \\
\hline
Romanian RRT (UDPipe) & \textbf{98.41} & 86.74 \\
Romanian RRT (DPSG, mT5) & - & 88.76\\
Romanian RRT (Ours, ByT5) & 98.15 & \textbf{91.16} \\
\hline
Bulgarian BTB (UDPipe) & 97.94 & 90.35 \\ 
Bulgarian BTB (DPSG, mT5) & - & 93.92 \\ 
Bulgarian BTB (Ours, ByT5) & \textbf{98.51} & \textbf{94.11} 

%% file: input/results-ocr.tex
ByT5-Small & 2.98 & 23.19 \\
ByT5-Sanskrit & \textbf{2.69} & \textbf{20.03} \\

%% file: input/results-new-dataset.tex
S  & 76.09 & 83.71 & 87.21 & 84.61 & \textbf{88.21} \\
L & 68.27 & 77.99 & 82.05 & 79.88 & \textbf{83.96} \\
\hline
S+M & 49.94 & 60.93 & 71.50 & 63.86 & \textbf{74.38} \\
L+M & 49.23 & 59.28 & 69.40 & 62.00 & \textbf{72.33} \\

\hline 
S+L+M & 49.10 & 58.75 & 71.92 & 61.27  & \textbf{74.31} 

%% file: input/ablation.tex
Segmentation only & 83.52 (-1.09) \\
Lemmatization only & 77.52 (-2.35)\\
\hline
Segmentation only w/o rec. & 81.58 (-3.03)\\
Lemmatization only w/o rec. & 76.85 (-3.03)